\documentclass{clv3}
\usepackage{pgfplots}
\pgfplotsset{%
    ,compat=1.12
    ,every axis x label/.style={at={(current axis.right of origin)},anchor=north west}
    ,every axis y label/.style={at={(current axis.above origin)},anchor=north east}
    }
\usepackage{hyperref}
\usepackage{xcolor}
\definecolor{darkblue}{rgb}{0, 0, 0.5}
\hypersetup{colorlinks=true,citecolor=darkblue, linkcolor=darkblue, urlcolor=darkblue}
\usepackage{algorithm}
\usepackage{algpseudocode}
\usepackage{amssymb}
\usepackage{rotating}
\usepackage[export]{adjustbox}
\usepackage{comment}
\usepackage{gb4e}
\usepackage{textcomp}
\usepackage{multirow}
\exewidth{(\thexnumi)}
\noautomath

\usepackage{tikz-dependency}
\tikzset{/depgraph/.cd,/depgraph/.search also = {/tikz},
baseline=-0.6ex, inner sep=-0.1cm, edge horizontal padding=3pt, edge unit distance=1.8ex}
\usepackage{array}

\issue{1}{1}{2020}

\runningtitle{A Hybrid Approach to Dependency Parsing}

\runningauthor{\c{S}aziye Bet\"{u}l \"{O}zate\c{s}}

\begin{document}

\title{A Hybrid Approach to Dependency Parsing: Combining Rules and Morphology with Deep Learning }

\author{\c{S}aziye Bet\"{u}l \"{O}zate\c{s}}
\affil{Department of Computer Engineering Bo\u{g}azi\c{c}i University  {\fontfamily{lmtt}\selectfont saziye.bilgin@boun.edu.tr} }

\author{Arzucan \"{O}zg\"{u}r}
\affil{Department of Computer Engineering Bo\u{g}azi\c{c}i University {\fontfamily{lmtt}\selectfont arzucan.ozgur@boun.edu.tr}}

\author{Tunga G\"{u}ng\"{o}r}
\affil{Department of Computer Engineering Bo\u{g}azi\c{c}i University {\fontfamily{lmtt}\selectfont gungort@boun.edu.tr}}

\author{Balk{\i}z \"{O}zt\"{u}rk}
\affil{Department of Linguistics\\ Bo\u{g}azi\c{c}i University \\
{\fontfamily{lmtt}\selectfont balkiz.ozturk@boun.edu.tr}}

\maketitle

\begin{abstract}
Fully data-driven, deep learning-based models are usually designed as language-independent and have been shown to be successful for many natural language processing tasks. However, when the studied language is low-resourced and the amount of training data is insufficient, these models can benefit from the integration of natural language grammar-based information.
We propose two approaches to dependency parsing especially for languages with restricted amount of training data. Our first approach combines a state-of-the-art deep learning-based parser with a rule-based approach and the second one incorporates morphological information into the parser. In the rule-based approach, the parsing decisions made by the rules are encoded and concatenated with the vector representations of the input words as additional information to the deep network. The morphology-based approach proposes different methods to include the morphological structure of words into the parser network. Experiments are conducted on the IMST-UD Treebank and the results suggest that integration of explicit knowledge about the target language to a neural parser through a rule-based parsing system and morphological analysis leads to more accurate annotations and hence, increases the parsing performance in terms of attachment scores. The proposed methods are developed for Turkish, but can be adapted to other languages as well.
\end{abstract}

\section{Introduction}

Current state-of-the-art dependency parsers usually rely solely on deep learning methods, where parsers try to learn the characteristics of the language from available training data \cite{dyer-etal-2015-transition,dozat2017deep}. As expected, this approach works well when the training data size is big enough. However, these pure deep learning-based approaches cannot reach the desired success levels when the data size is insufficient \cite{srivastava2014dropout}. It is observed that deep learning-based systems need large amounts of data to be able to reach high performance \cite{sudha2015survey}. For languages with small data sets, there is a need for developing additional methods that meet the characteristic needs of these languages. 

In this article, we propose to take into account the language grammar and integrate the information extracted from the grammar to a deep learning-based dependency parser. We propose two approaches for the inclusion of the grammar to the neural parser model. Our first approach is to integrate linguistically-oriented rules to a deep learning-based parser for dependency parsing of languages especially with restricted amount of training data. The rules are created to deal with the problematic parts in the sentences that are hard to predict. In our second approach, we give morpheme information as an additional input source to the parsing system. We experimented with different methods for inclusion of the morpheme information. We applied the proposed methods to Turkish and the experimental results suggest that both approaches improve the parsing performance of a state-of-the-art dependency parser for Turkish.

The proposed methods were evaluated on both projective and non-projective sentences and  currently hold the state-of-the-art performance in parsing the Turkish IMST-UD Treebank. To the best of our knowledge, this is the first study that integrates the parsing actions of a rule-based method and morphological elements into a deep learning-based dependency parsing system and may serve as a base for other low-resource languages. 

The main contributions of this article are as follows:
\begin{itemize}
    \item A novel rule-based enhancement method that can be integrated to any neural dependency parser.
    \item A morphology-based enhancement method with three different ways of including morphological information to the parser.
    \item A simple yet useful integration method that allows to combine the proposed enhancement methods with any neural dependency parser.
    \item State-of-the-art dependency parsing scores on the IMST-UD Treebank.
\end{itemize}

The rest of this article is organized as follows. Section 2 presents the related work on deep learning-based, rule-based, and morphology-based approaches to dependency parsing. It also describes the related work on Turkish dependency parsing. In Section 3, we describe our proposed models to dependency parsing that combine hand-crafted rules and the morphological information with a state-of-the-art deep learning-based dependency parser. Section 4 explains how these models can be adapted to other languages. Section 5 gives the experiment details and results. Finally, Section 6 concludes the article and suggests some future work.

\section{Related Work}

Purely rule-based approaches to NLP problems have been very popular in the past, from part of speech tagging \cite{brill1992simple} to aspect extraction in sentiment analysis \cite{poria2014rule}. Rule-based methods have also been applied to dependency parsing. There have been studies on rule-based parsing using grammar rules for Turkish \cite{oflazer2003dependency} and for other languages \cite{sennrich2009new,ramasamy2011tamil,boguslavsky2011rule,korzeniowski2017rule}.  

Recently, deep learning methods are frequently applied to dependency parsing and show promising performances in predicting the dependency parses of sentences \cite{dyer-etal-2015-transition,kiperwasser2016simple,dozat2017deep}. In 2017, a state-of-the-art LSTM-based dependency parser \cite{dozat2017stanford} achieved the best performance in 54 treebanks including the IMST-UD Treebank at the CoNLL'17  Shared Task UD Parsing \cite{zeman-EtAl:2017:K17-3}. This parser together with its enhanced versions \cite{kanerva-EtAl:2018:K18-2,che-EtAl:2018:K18-2} presented at the CoNLL'18 Shared Task on UD Parsing \cite{zeman-EtAl:2018:K18-2} currently hold the state-of-the-art performances on the dependency parsing of many languages. However, these parsers do not have language specific features that can boost the parsing performance, especially for morphologically rich and under-resourced languages like Turkish. 

Morphologically rich languages (MRLs) pose problems when state-of-the-art NLP models developed for the most widely studied languages like English and French are applied directly to them \cite{tsarfaty2010statistical}.
There are studies that include rule-based knowledge to data-driven parsers in order to increase parsing accuracy. \citet{zeman2005improving} experimented with different voting mechanisms to combine seven different dependency parsers including a rule-based parser. Another study applied a rule-based mechanism on the output of a dependency parser to create collapsed dependencies \cite{ruppert2015rule}. 

There have also been several approaches that use morphological information in the dependency parsing of the MRLs. Similar to \citet{ambati2010role} and \citet{Goldberg:2010:EFD:1868771.1868783}, which utilize morphological features for the dependency parsing of Hindi and Hebrew respectively,  \citet{marton2010improving} measured the effects of nine morphological features extracted from an Arabic morphological analysis and disambiguation toolkit \cite{habash2005arabic} on the parsing performance of the MaltParser \cite{nivre2007maltparser} for Arabic. These studies show that usage of some morphological features works well for the dependency parsing of MRLs. \citet{vania-etal-2018-character} compared the strength of character-level modelling of words with an oracle model which has explicit access to morphological analysis of words on dependency parsing and observed that combining words with their corresponding morpheme information using a bi-LSTM structure in the word representation layer outperforms the character-based word representation models.
\citet{ozatecs2018morphology} proposed two morphology-based approaches to dependency parsing. In their first approach, they combined the vector representation of words with the representation of some of the morphological attributes given in treebanks. Their second approach represents the words by separating them to their corresponding lemma and suffixes for suitable languages. They observed that both of the models improve the parsing accuracy for agglutinative languages. 
\citet{dehouck2019phylogenetic} proposed a multi-task learning framework that makes use of language philogenetic trees to represent the shared information among the languages. They used gold morphological features for dependency parsing by summing the created vectors of each morphological attribute given in the treebanks and add this vector to the representation of the word, similarly to \citet{ozatecs2018morphology}.

However, to the best of our knowledge, there does not exist any prior research on a hybrid approach for dependency parsing, where parsing decisions of hand-crafted rules together with morphological information are integrated into a deep learning-based dependency parsing model. Our inclusion of morphology also differs from the previous works in terms of extracting the morphological information. Instead of using morphological features, our models utilize suffixes of a word explicitly. While two of the proposed morphology-based methods include the suffixes of each word to the word representation model directly, the third one represents each word with the suffixes which can and cannot bind to the root of that word. 

\subsection{Turkish Dependency Parsing}
In this study, we propose both rule-based and morphology-based enhancement methods for dependency parsing and integrate our methods to a state-of-the-art dependency parser \cite{dozat2017stanford}. We applied the proposed approaches to Turkish. Because unlike English, the resources for Turkish natural language processing (NLP) in general are very restricted. For dependency parsing, the English treebanks have a total of 34,631 sentences annotated in the Universal Dependencies (UD) style \cite{ud}, with the largest one, the EWT Treebank \cite{silveira2014gold} including 16,622 sentences. On the other hand, for Turkish, the only data sets used for training and evaluation of the systems are the IMST-UD Treebank \cite{sulubacak-etal-2016-universal} which consists of 5,635 annotated sentences and the Turkish UD Parallel (PUD) Treebank \cite{zeman-EtAl:2017:K17-3} of 1,000 annotated sentences that is used for testing purposes. There is also IWT-UD Treebank \cite{sulubacak2018implementing}, which includes 5,009 sentences crawled from the web. However the sentences in this treebank are from social media texts and the treebank has a non-canonical language that is completely different from the other well-edited treebanks \cite{sulubacak2018implementing}.

Besides being a low-resourced language, Turkish is not a well-studied language in natural language processing, and dependency parsing is no exception to this.  Following the initial work in \citet{oflazer2003dependency}, another study presents a word-based and two inflectional group-based input representation models for dependency parsing of Turkish \cite{eryiugit2006statistical} which use a version of backward beam search to parse the sentences. They used a subset of 3,398 sentences of the Turkish Dependency Treebank \cite{oflazer2003building} with only projective (non-crossing) dependency relations to train and test the proposed parsers. Later, a data-driven dependency parser for Turkish was proposed, which employs no grammar, but relies completely on inductive learning from treebank data for the analysis of new sentences, and on deterministic parsing for disambiguation \cite{eryiugit2008dependency}. The authors use a variant of the parsing system MaltParser proposed in \citet{nivre2007maltparser}, a linear-time, deterministic, classifier-based parsing method with history-based feature
models and discriminative learning. Only the projective dependency relations in the treebank were used for evaluation. 

\citet{eryiugit2011multiword} extracted different multi-word expression classes as a pre-processing step for a statistical dependency parser. Only the projective dependencies are considered. 
\citet{hall2010single} and \citet{el2014initial} are other notable studies that are based on optimized versions of the MaltParser system. 
Later, a graph-based approach was proposed in \citet{seeker2015graph}, where a discriminative linear model is trained and a lattice dependency parser is created that uses dual decomposition. All these studies on Turkish dependency parsing used the first version \cite{oflazer2003building} of the Turkish Treebank annotated in non-UD (non-Universal Dependencies) style. 

To eliminate the inconsistencies in this treebank and to obtain better parsing results, a revised version of it is presented under the name of IMST Treebank \cite{sulubacak2016imst} and this new version is evaluated using the MaltParser without the non-projective dependencies. However, the IMST Treebank is also in non-UD style. To contribute to the unifying efforts of the UD project, the IMST Treebank is converted automatically to the UD annotation style \cite{sulubacak-etal-2016-universal} and is named as IMST-UD Treebank. This most up-to-date version of the Turkish treebank is evaluated using MaltParser in \citet{sulubacak2018implementing}, however only a subset of the treebank is used in the evaluation by eliminating the non-projective dependencies. In our study, we include both projective and non-projective dependencies in the IMST-UD Treebank as the proportion of non-projective sentences in Turkish is too high (over 20\%) to be ignored \cite{hall2010single}.

\section{The Proposed Method}

In order to improve the parsing performance of deep learning-based parsers, we design hybrid methods where the grammar-based information is fed into the deep network of a data-driven parser. We propose two different approaches for supplying the information extracted from the language grammar, the first one is via hand-crafted grammar rules for detecting dependency relations between words and the second one is by analyzing the underlying morphological structure of words.

We first give a brief description of the state-of-the-art neural parser used in this study. We then explain our rule-based parsing method and show how these two methods are integrated to get a better parsing mechanism. Finally, we describe our morphology-based enhancement method and its integration to the parser.

\begin{figure}[h]
\centering
 \includegraphics[width=0.8\textwidth]{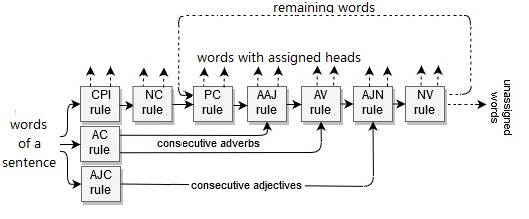}
                \caption{ Our rule-based dependency parser. } 
\label{rule-based-parser}
\end{figure}

\subsection{Stanford's Neural Dependency Parser}

Stanford's graph-based neural dependency parser \cite{dozat2017stanford} is the leading system in the CoNLL'17 Shared Task on UD Parsing \cite{zeman-EtAl:2017:K17-3}. It uses unidirectional LSTM modules to create word embeddings, which consist of learned word embeddings,  pre-trained word embeddings, and character embeddings. These word embeddings are then concatenated with their corresponding part-of-speech (POS) tag embeddings and given to the bidirectional LSTM modules. They use ReLu layers and biaffine classifiers to score possible head and dependency relations. 
For more information, see \cite{dozat2017stanford}. In this study, we use this parser as the baseline system. 

\subsection{A Rule-based Unlabeled Parsing Approach}
\label{sec:rulebased}
We designed linguistically oriented rules for the dependency parsing of the Turkish language. These rules determine the head of the words in a sentence without assigning a label for the created dependency relation. Rather than a complete parser that creates a fully connected dependency graph, our system deals with the most difficult cases in predicting dependency relations in a sentence according to the parsing errors, such as complex predicates or multiple adverbs. The rules are created by considering the relations between the main elements of a sentence. We consider the relations between verbs, nouns, adverbs, and adjectives in a sentence as having the main importance and generate rules that deal with these relations. The rules are based on the existing grammar rules extracted form \cite{goksel2005}. 

Figure \ref{rule-based-parser} shows the general mechanism of our rule-based parsing system. Our model takes a tokenized sentence as its input. First, the AC rule creates a list of the consecutive adverbs that are related to each other in the sentence, and sends this list to the AAJ and AV rules as an input. Similarly, the AJC rule creates a list of the consecutive adjectives in the sentence that are related to each other and sends this list to the AJN rule. After that, the CPI and NC rules are applied to the words of the sentence sequentially. Then, the PC, AAJ, AV, AJN, and NV rules are applied to the remaining words in an iterative manner. This process continues until the heads of all the words in the sentence are associated with a dependency relation or no more dependency relations can be found in the sentence. The following subsections explain each rule in detail.

\subsubsection{Complex Predicates and Verbal Idioms (CPI) Rule}
This rule finds the complex predicates (e.g., \textit{kabul et (accept)}, \textit{sebep ol (cause)} etc.) and verbal idioms (e.g., \textit{g{\"o}z yum (condone)} etc.) in Turkish. Complex predicates are made up of a bare nominal followed by one of the free auxiliary verbs \textit{ol, et, yap, gel, dur, kal, çık, düş, buyur, eyle} \citep[][page 143]{goksel2005}.

However, verbal idioms can have verbs in a wide range of words and the meaning of these verbs are changed when they are used in an idiom. Due to insufficient amount of training data, parsers usually fail to detect these multi-word predicates \cite{ramisch-etal-2018-edition} and consider the verb of such predicates as the head of the relation. Example \ref{compound_verb} shows such a false annotation done by the trained baseline deep learning-based parser.\footnote{False annotations are shown with dotted lines, their corrected forms are shown with fine lines. Thick lines represent the annotations predicted correctly by the parser. This representation is followed in all of the examples in this article.} The parser falsely predicts that the verb \textit{getir (bring)} is the \texttt{root} word and \textit{yerine (to its place)} is an \texttt{oblique} of the \texttt{root} word. In fact, \textit{yerine getir (fulfill)} is a verbal idiom and the verb \textit{getir} is the verbal component of the complex predicate.
\begin{exe}
    \ex
        \scalebox{1}{
        \begin{dependency}
            \begin{deptext}[column sep=.3cm]
            Her \& istedi\u{g}ini \& yerine \& getiriyordum \\
        \end{deptext}
            \deproot[edge unit distance=2.5ex]{3}{\textsc{root}}
            \deproot[edge below, edge unit distance=2.5ex,edge style={densely dotted}]{4}{\textsc{root}}
            \depedge[edge style={ultra thick}]{2}{1}{\textsc{\textbf{det}}}
            \depedge{3}{4}{\textsc{compound}}
            \depedge[edge below, edge unit distance= 2ex,edge style={densely dotted}]{4}{3}{\textsc{obl}}
            \depedge{3}{2}{\textsc{obj}}
            \depedge[edge below, edge style={densely dotted}]{4}{2}{\textsc{obj}}

        \end{dependency}
        }
    \gll Her \hspace{12pt} iste-di-\u{g}i-ni yer-in-e getir-iyor-dum. \\
    every \hspace{12pt} want-\textsc{pst}-\textsc{3sg.poss}-\textsc{acc} place-\textsc{poss}-\textsc{dat} bring-\textsc{prs}-\textsc{pst}-\textsc{1sg} \\
    \glt{`I have been doing whatever he/she wants.'}
    \label{compound_verb}
\end{exe}

This rule correctly constructs the dependency relations between the words of such predicates. In order to detect such verbal compounds however, it needs a dictionary that lists complex predicates and idioms in Turkish. We collected approximately 8K complex predicates using the Turkish Proverbs and Idioms Dictionary supplied by the Turkish Language Association (TDK) \cite{TDKweb} and from various online Turkish resources.\footnote{All of the lexicons collected for this study as well as the source codes of the proposed models will be made publicly available upon publication of the paper.} The CPI rule searches for complex predicates and idioms in a sentence. When such a predicate is found, the second word of the predicate is set as a dependent of the first word. Since the head of the second word is found, it is eliminated from the remaining word list. The first word whose type is noun is marked as {\it CP} to denote that it is not a simple noun but a part of a complex predicate. 

\subsubsection{Noun Compounds (NC) Rule}
In Turkish, there are two types of noun compounds: bare compounds that are treated as head-level ($X^0$) constructions and -$(s)I(n)$ compounds where the first noun has no suffixes while the second noun in the compound takes the 3rd person possessive suffix -$(s)I(n)$ \citep[][page 94]{goksel2005}. In terms of dependency grammar representations in ($X^0$) constructions the head word of the compound is the first noun, whereas in -$(s)I(n)$ compounds the first noun is dependent on the second noun. Differentiating between these two noun compound types is not easy for parsers in the absence of large amount of training data. Example \ref{nc1} shows an ($X^0$) compound \textit{kuru yemiş (dried nuts)}. According to the generative grammar tradition, the word that relates this compound structure to the rest of the sentence should be \textit{yemiş (nuts)} \cite{goksel2007remarks}. However, this is not the case in the dependency grammar representations \cite{de2019dependency}. So, in Example \ref{nc1}, both of the words are in their bare forms with no suffixes and form a noun compound with the head word being \textit{kuru (dried)}. Yet, the parser falsely predicts \textit{kuru} as an adjective modifier of \textit{yemi\c{s} (nuts)}.

\begin{exe}
    \ex 
    \begin{varwidth}{\linewidth}
        \scalebox{1}{
        \begin{dependency}
            \begin{deptext}[column sep=.3cm]
            Kuru \& \& yemi\c{s}  \\
        \end{deptext}
            \depedge[edge below, edge style={densely dotted}]{3}{1}{\textsc{amod}}
            \depedge{1}{3}{\textsc{compound}}
        \end{dependency}
        }
    \\
    \glt `Dried \hspace{16pt} nuts'
    \label{nc1} \end{varwidth} \hfil
\end{exe}

The second type of compounds, \textit{-(s)I(n)} compounds, can easily be confused with a noun followed by another noun with the accusative suffix \textit{-(y)I} that do not form a compound. (A more detailed discussion about this confusion is given in Section \ref{PC} accompanied with examples \ref{nom} and \ref{acc}.) 

In addition to these two types of noun compounds, reduplicated compounds are also handled by this rule. Reduplication is a common process in Turkish \cite{erguvanli1984function}. Reduplicated words construct reduplicated compounds. However, the parser sometimes cannot recognize this idiomatic structure and fails to construct the compound. Example \ref{nc2} shows this kind of confusion where the parser falsely assigns the first word of the compound as an adjective modifier of the second.

\begin{exe}
    \ex  
    \begin{varwidth}{\linewidth}
        \scalebox{1}{
        \begin{dependency}
            \begin{deptext}[column sep=.3cm]
            Arka \& \& arkaya  \\
        \end{deptext}
            \depedge[edge below, edge style={densely dotted}]{3}{1}{\textsc{amod}}
            \depedge{1}{3}{\textsc{compound:redup}}
        \end{dependency}
        }
   \gll arka arka-ya \\
        back back-DAT \\
    \glt{`Repeatedly'}
    \label{nc2} \end{varwidth} \hfil
\end{exe}

To separate between these three cases, the NC rule utilizes large lexicons and detects the noun compounds in sentences with the help of these lexicons. We extracted three different lexicons from the official noun compounds dictionary of Turkish language published by the Turkish Language Association. These three lexicons are for noun compounds, possessive compounds, and reduplicated compounds with, respectively, the sizes of approximately 5K, 6K, and 2K entries. We classified each entry in the dictionary as one of the three kinds of compounds according to their lexical classes. The NC rule searches through these lexicons and by this way detects noun compounds, possessive compounds, and reduplicated compounds in the sentences.

\subsubsection{Possessive Construction (PC) Rule}
\label{PC}
The PC rule includes both \textit{genitive possessive constructions} and \textit{possessive compounds} that cannot be detected by the NC rule. In genitive possessive constructions, the first noun of a noun phrase takes a genitive suffix \textit{-(n)In} that represents the ownership of the first noun over the second noun. The second noun takes a possessive suffix \textit{-(s)I(n)} \citep[][page 161]{goksel2005}. In possessive compounds, there is no genitive suffix and the first noun in the compound appears without any case marking \citep[][page 96]{goksel2005}. Although it is easy for the parser to detect the genitive possessive construction relations due to the existence of a genitive suffix, detecting possessive compounds is a challenging task. Because in possessive compounds, there does not exist a genitive suffix in the first noun and the possessive suffix \textit{-(s)I(n)} in the second noun is confusing since it appears the same as the accusative suffix \textit{-(y)I} when \textit{(s)} is dropped. 

This situation is depicted in Examples \ref{gen}, \ref{nom} and \ref{acc}. The only difference in Examples \ref{gen} and \ref{nom} is that the genitive suffix showing the possession exists in \ref{gen} and is omitted in \ref{nom}. In both sentences, the subject of the sentence is the word \textit{ya\u{g} (oil)} and the two nouns form a possessive construction. However, this is not the case in Example \ref{acc}. Here, the subject of the sentence is the word \textit{makine (machine)}, and the word \textit{ya\u{g} (oil)} is its object. The confusion originates from the use of the same consecutive nouns, \textit{makine ya\u{g}{\i}} in both of the example sentences \ref{nom} and \ref{acc}. However, the \textit{-{\i}} suffix of the word \textit{ya\u{g}} in \ref{acc} is actually an accusative suffix and hence the two nouns in \ref{acc} do not form a compound. In order to help the parser to differentiate between these two cases in sentences, we construct the PC rule that identifies whether there is a compound relation between two consecutive nouns or not. 

When two consecutive nouns are detected, the rule checks whether there is a genitive suffix in the first noun. If yes, it is set as a dependent of the second noun. If the first noun is in bare form and the second noun has a possessive suffix, then the first noun is set as the dependent on the second noun. As stated, the third person possessive suffix {\it -(s)I(n)} can be confused with the accusative suffix {\it -(y)I} when they are attached to a word ending in a consonant. In this case, both suffixes reduce to the form \textit{{\i}, i, u,} or \textit{\"{u}}. To prevent this confusion, the rule analyzes the morphological features of the word and checks if it is identified as accusative (example \ref{acc}) or not (example \ref{nom}).
\begin{exe}
    \ex  
    \begin{varwidth}{\linewidth}
        \scalebox{1}{
        \begin{dependency}
            \begin{deptext}[column sep=.3cm]
            Makinenin \& ya\u{g}{\i} \& akt{\i} \\
        \end{deptext}
            \deproot[edge unit distance=2ex]{3}{\textsc{root}}
            \depedge{2}{1}{\textsc{\textbf{nmod:poss}}}
            \depedge{3}{2}{\textsc{nsubj}}

        \end{dependency}
        }
    \gll Makinenin ya\u{g}-{\i} ak-t{\i}. \\
    machine-\textsc{gen} oil-\textsc{3sg}-\textsc{poss} leak-\textsc{pst}-\textsc{3sg} \\
    \glt{`The machine's oil leaked.'}
    \label{gen} \end{varwidth} \hfil
\end{exe}

\begin{exe}
    \ex  
    \begin{varwidth}{\linewidth}
        \scalebox{1}{
        \begin{dependency}
            \begin{deptext}[column sep=.3cm]
            Makine \& ya\u{g}{\i} \& akt{\i} \\
        \end{deptext}
            \deproot[edge unit distance=2ex]{3}{\textsc{root}}
            \depedge{2}{1}{\textsc{\textbf{nmod}}}
            \depedge{3}{2}{\textsc{nsubj}}

        \end{dependency}
        }
    \gll Makine ya\u{g}-{\i} ak-t{\i}. \\
    machine oil-\textsc{3sg}-\textsc{poss} leak-\textsc{pst}-\textsc{3sg} \\
    \glt{`Machine oil leaked.'}
    \label{nom}  \end{varwidth} \hfil
\end{exe}
\begin{exe}
    \ex 
    \begin{varwidth}{\linewidth}
        \scalebox{1}{
        \begin{dependency}
            \begin{deptext}[column sep=.3cm]
            Makine \& ya\u{g}{\i} \& ak{\i}tt{\i} \\
        \end{deptext}
            \deproot[edge unit distance=2ex]{3}{\textsc{root}}
            \depedge{3}{1}{\textsc{\textbf{nsubj}}}
            \depedge{3}{2}{\textsc{obj}}

        \end{dependency}
        }
    \gll Makine  ya\u{g}-{\i} ak{\i}t-t{\i}. \\
    machine oil-\textsc{acc} drip-\textsc{pst}-\textsc{3sg} \\
    \glt{`The machine dripped the oil.'}
    \label{acc}
     \end{varwidth}\hfil
\end{exe}
In addition, the PC rule deals with multi-word proper nouns and determiner-noun relations. When the PC rule detects a multi-word proper noun, all the following consecutive proper nouns are set as dependent on the first proper noun. When it detects a noun that is preceded by a determiner, it sets the determiner as dependent on the noun.

\subsubsection{Consecutive Adverb (AC) Rule}

Consecutive adverbs are also hard-to-detect predicates with data-driven parsers. For instance, the first adverb \textit{sonra (then)} and the second adverb \textit{\c{c}ok (very)} are both dependents of the verb \textit{\c{s}a\c{s}{\i}r{\i}rs{\i}n{\i}z (be surprised)} in Example \ref{compound_adv}. However, the parser falsely predicts the word \textit{sonra} as the dependent of the previous word \textit{inan{\i}rsan{\i}z (if you believe)} with \texttt{case} label. 
The AC rule handles such consecutive adverbs in a sentence. We observe that, if there are two consecutive adverbs in a sentence, usually there are two cases: either the first adverb is dependent on the second adverb or they are both dependent on the same head word. So, when two consecutive adverbs are found, the method checks whether the first adverb belongs to the group of adverbs that emphasize the meaning of the next adverb or not \citep[][page 213]{goksel2005}. This is done via searching through a list of adverbs of quantity or degree taken from \citep[][pages 210-211]{goksel2005}. If yes, the first adverb is set as a dependent word of the second adverb and the first adverb is dropped. If not, these two adverbs are put in a list (which will be called as {\it the consecutive adverbs list} throughout the article) and the first one is dropped from the list of remaining words. When the head word of the second adverb is found later, the first adverb is also bound to the same head word. 

\begin{exe}
    \ex
        \scalebox{1}{
        \begin{dependency}
            \begin{deptext}[column sep=.3cm]
            \.{I}nan{\i}rsan{\i}z \& sonra \& \c{c}ok \& \c{s}a\c{s}{\i}r{\i}rs{\i}n{\i}z  \\
        \end{deptext}
            \deproot[edge unit distance=2.5ex,edge style={ultra thick}]{4}{\textsc{root}}
            \depedge[edge style={ultra thick}]{4}{3}{\textsc{advmod}}
             \depedge[edge below,edge style={densely dotted}]{1}{2}{\textsc{case}}
              \depedge{4}{1}{\textsc{advcl:cond}}
               \depedge[edge below,edge unit distance=1.1ex, edge style={densely dotted}]{4}{1}{\textsc{nmod}}
               \depedge{4}{2}{\textsc{advmod}}

        \end{dependency}
        }
    \gll  \.{I}nan-{\i}r-sa-n{\i}z sonra \c{c}ok \c{s}a\c{s}{\i}r-{\i}r-s{\i}n{\i}z. \\
    believe-\textsc{aor}-\textsc{cond}-\textsc{2pl} then very surprise-\textsc{aor}-\textsc{2pl} \\
    \glt{`If you believe, then you will be very surprised.'}
    \label{compound_adv}
\end{exe}

\subsubsection{Consecutive Adjective (AJC) Rule}
Consecutive adjectives are another troublesome word group which the parser sometimes fails to parse correctly. Example \ref{compound_adj} shows such an annotation.
\begin{exe}
\ex 
    \begin{varwidth}{\linewidth}
        \scalebox{0.9}{
        \begin{dependency}
            \begin{deptext}[column sep=.3cm]
            Asistan{\i}m \& bulan{\i}k \& anlams{\i}z \& g\"{o}zlerini \& bana \& \c{c}evirdi \\
        \end{deptext}
            \deproot[edge unit distance=3.5ex,edge style={ultra thick}]{6}{\textsc{root}}
           \depedge[edge style={ultra thick}]{6}{4}{\textsc{obj}}
            \depedge[edge style={ultra thick}]{6}{1}{\textsc{nsubj}}
            \depedge{6}{5}{\textsc{iobj}}
             \depedge[edge below,edge style={densely dotted}]{6}{5}{\textsc{obj}}
              \depedge[edge unit distance=1ex,edge below,edge style={densely dotted}]{6}{2}{\textsc{nsubj}}
              \depedge[edge style={ultra thick}]{4}{3}{\textsc{amod}}
               \depedge{4}{2}{\textsc{amod}}

        \end{dependency}
        }
    \gll Asistan{\i}m bulan{\i}k anlams{\i}z g\"{o}z-ler-in-i bana \c{c}evir-di. \\
    assistant-\textsc{1sg.poss} blurred meaningless eye-\textsc{pl}-\textsc{3sg.poss}-\textsc{acc} I-\textsc{dat} turn-\textsc{pst}-\textsc{3sg} \\
    \glt{`My assistant turned his/her blurred meaningless eyes towards me.'}
    \label{compound_adj}
    \end{varwidth}\hfil
\end{exe}

The parser falsely considers the word \textit{bulan{\i}k (blurred)} as the subject of the sentence and assigns \texttt{nsubj} to \textit{bulan{\i}k}. In fact, it is an adjective describing the word \textit{g\"{o}zler (eyes)} and should be an \texttt{amod} dependent of the word \textit{g\"{o}zler}.
The AJC rule is created to prevent this type of errors. 
This rule finds all the consecutive adjectives in a sentence. Usually, two consecutive adjectives are dependent on the same word. So, when two consecutive adjectives are found, these adjectives are put into a list (which will be called as {\it the consecutive adjectives list} from now on) and the first one is dropped from the list of remaining words. When the head word of the second adjective is found later by the parser, the first adjective is also set as a dependent of the same head word.

\subsubsection{Adverb-Adjective (AAJ) Rule}
The AAJ rule handles adverb-adjective relations in a sentence.
For every two consecutive words in the sentence, if the first word is an adverb and the second word is an adjective, and if the adverb is a quantity or degree adverb,
then the adverb is set as a dependent of the adjective word \citep[][pages 175-180]{goksel2005}. When the head of an adverb is obtained in this way, the rule checks whether the adverb is in {\it the consecutive adverbs list} supplied by the AC rule. If yes, the consecutive adverbs of this adverb are also set as dependents of the same head word.

\subsubsection{Adverb-Verb (AV) Rule}

 For every two consecutive words in a sentence, if the first word is an adverb and the second word is a verb, and if the adverb is not one of {\it bile (even), -DAn \"{o}nce (before something), -DAn sonra (after something)} that emphasize the preceeding word, then the AV rule sets the adverb as a dependent of the verb \citep[][page 189]{goksel2005}. Otherwise, it sets the previous word of the adverb as its head. As the head of an adverb is found, the AV rule checks whether the adverb is in  {\it the consecutive adverbs list} supplied by the AC rule. If yes, the same head word is assigned to its compound adverbs. 

\subsubsection{Adjective-Noun (AJN) Rule}

The AJN rule constructs adjective-noun relations.
For every two consecutive words in the sentence, if the first word is an adjective and the second word is a noun, then the adjective is set as a dependent word of the noun when they form an adjectival \citep[][page 170]{goksel2005}. Like for the adverbs, when the head of an adjective is found, the algorithm checks whether the adjective is in {\it the consecutive adjectives list} supplied by the AJC rule. If yes, the consecutive adjectives of this adjective are also set as dependents of the same head word. 

\subsubsection{Noun-Verb (NV) Rule}

\begin{figure}%[!h]
\centering
 \includegraphics[width=0.99\textwidth]{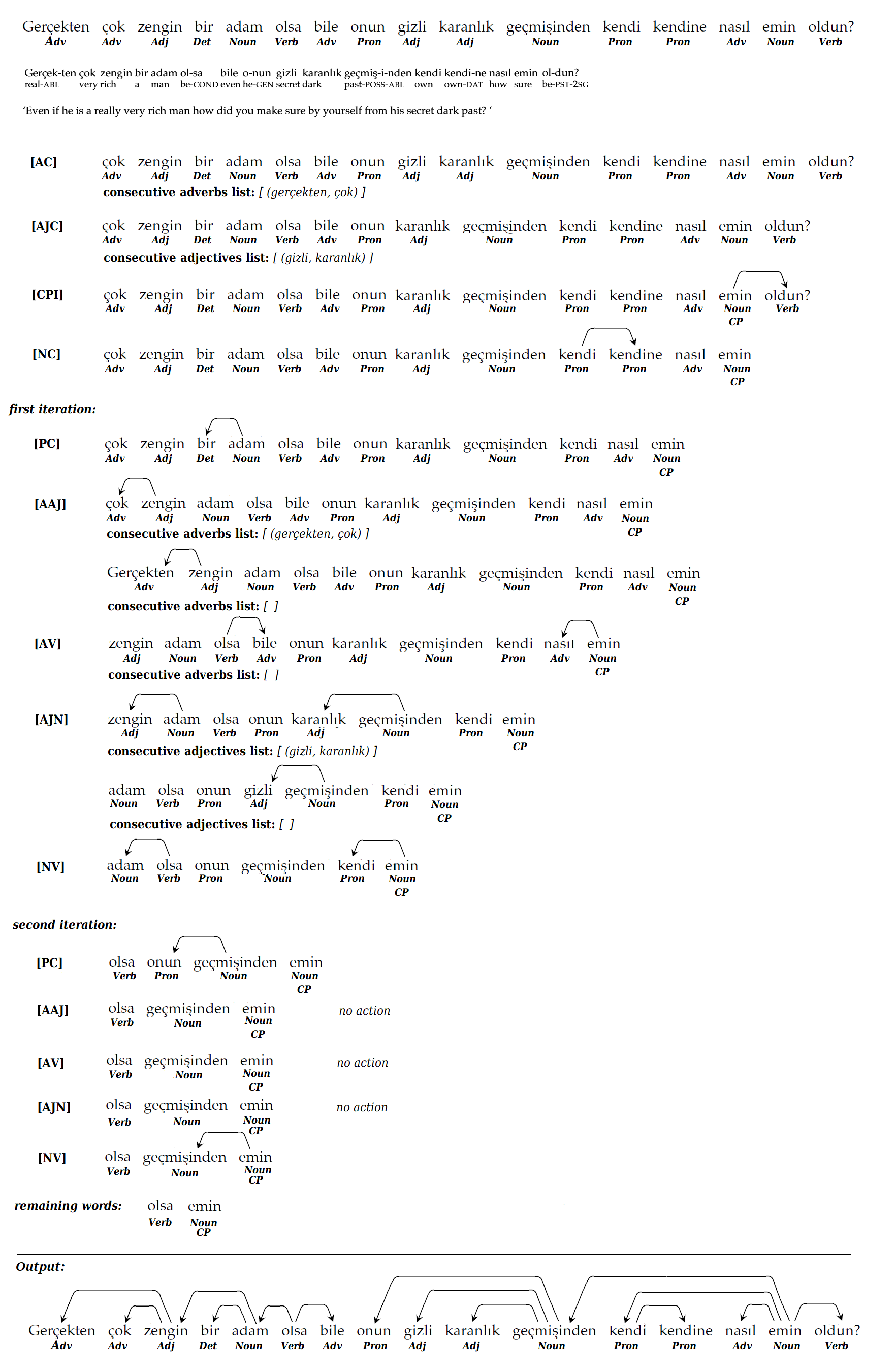}
                \caption{ The operation of the rule-based parser on an example sentence. } 
            
\label{bigexample}
\end{figure}

After complex predicates and noun, adverb, and adjective compounds are detected and eliminated from the sentence, the final NV rule assigns any unassigned noun or pronoun followed by a verb as a dependent of that verb.
\\
\\
Figure \ref{bigexample} depicts the application of each rule on an example sentence. In the example, the first rules that are applied to the sentence are the AC and AJC rules. These rules prepare {\it the consecutive adverbs list} and {\it the consecutive adjectives list}, respectively. {\it The consecutive adverbs list} stores the consecutive adverbs that should be bound to the same head word. Similarly, {\it the consecutive adjectives list} stores the consecutive adjectives which should have the same head word.

After that, the CPI and NC rules are applied consecutively. The CPI rule finds the complex predicates and idioms in the sentence using a large lexicon while the NC rule detects the noun compounds, possessive compounds, and reduplicated compounds that also exist in the pre-built lexicons.

As the operations of these one-time rules are completed, the PC, AAJ, AV, AJN, and NV rules are applied to the sentence in a loop until none of the rules can be applied anymore. As for the example sentence in Figure \ref{bigexample}, only two words remained unassigned out of sixteen.

\subsection{Integrating the Rule-based Approach with Stanford's Graph-based Parsing Method}
\label{intergration}

\begin{figure}[!h]
\centering
 \includegraphics[width=0.7\textwidth]{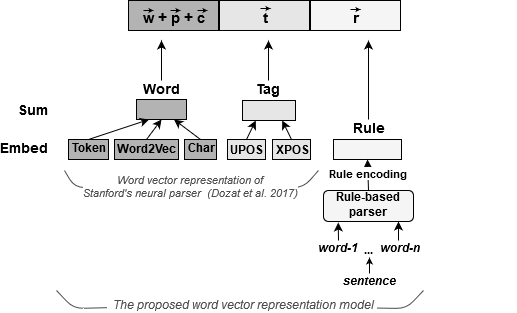}
                \caption{ Word embedding representation of the hybrid model with rule-based enhancement. } 
            
\label{rule-based-emb}
\end{figure}

Our approach for combining the rule-based method with Stanford's neural dependency parser is to embed the dependency parsing rule information to the dense representations of the words. The purpose of this approach is to give the parser an idea about finding the correct head of the corresponding word. 
The parser uses the dependent-head decisions made by the rule-based method in its learning phase and comes up with more accurate predictions about the syntactic annotation of sentences.

In our method, first the input sentences are pre-processed by the rule-based system. 
For each word in a sentence, the rules decide the head of the word if applicable, and then a three-letter encoding that denotes the rule action (CPI, PC, etc.) applied is assigned to that word. The rule-encoded words are dropped from the remaining words list and the rule-based system continues its process in a recursive manner until the head word of all words in the sentence are found or no more rule can be applied. Note that, each word is affected by at most one rule. After this rule-based pre-processing step, the rule encoding of each word is embedded with the same embedding technique used by the Stanford's parser \cite{dozat2017stanford}. Then this embedding vector of the rule encoding is concatenated to the embedding vector of that word, similar to the embedding vector of its POS-tag. Figure \ref{rule-based-emb} depicts this scheme.  

So, in our model, the rule vector representation is concatenated with the vector representation of \citet{dozat2017stanford}. The parser takes these word vector representations with the rule-based parsing decisions as input and learns the relations between the words in the training phase. In this way, we anticipate that the parser will benefit from the decisions of the rule-based system and arrive at a more accurate dependency tree for a given sentence.

\subsection{A Morphology-based Enhancement for Dependency Parsing}
\label{sec:morp}
In addition to the rule-based enhancement to the deep learning-based parser, we also propose to include the morphological information directly to the system. In this approach, we use morphemes of a word as an additional source. Our motivation relies on the fact that Turkish is a highly agglutinative language where the word structure is described by identifying the different categories of suffixes and determining which stems the suffixes may attach to and their orders. 
The suffixation process in Turkish sometimes produces very long word forms that correspond to whole sentences in English \cite{goksel2005}. The morphemes of a word hold some important information in terms of the dependency relations that word belongs to. For instance, it is observed that the last inflectional morpheme of a word determines its role as a dependent in the sentence \cite{oflazer2003building}. Based on this observation, we design three different methods to enhance the deep learning-based parser. The following subsections explain each model in detail.

\subsubsection{The Inflectional Suffixes Model}

\begin{figure}[!h]
\centering
 \includegraphics[width=0.7\textwidth]{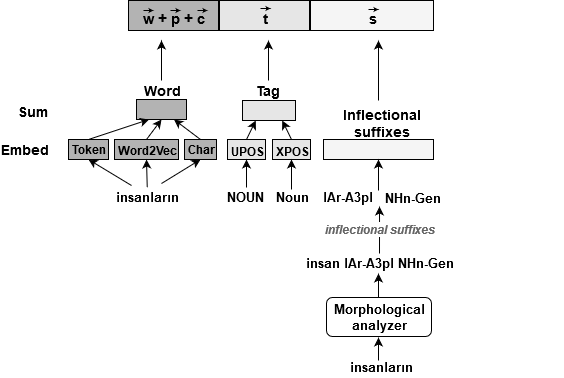}
                \caption{ Dense representation of the word \textit{insanlar{\i}n (people's)} using the Inflectional Suffixes Model. } 
            
\label{infsuffix}
\end{figure}

In this model, all of the inflectional suffixes are extracted from the morphologically analyzed version of the word, embedded, and then concatenated to the vector representation of that word. The integration method is the same as in Section \ref{intergration}. Figure \ref{infsuffix} depicts the creation of the dense representation of an example word \textit{insanlar{\i}n (people's)}.

\subsubsection{The Last Suffix Model}

\begin{figure}[!h]
\centering
 \includegraphics[width=0.7\textwidth]{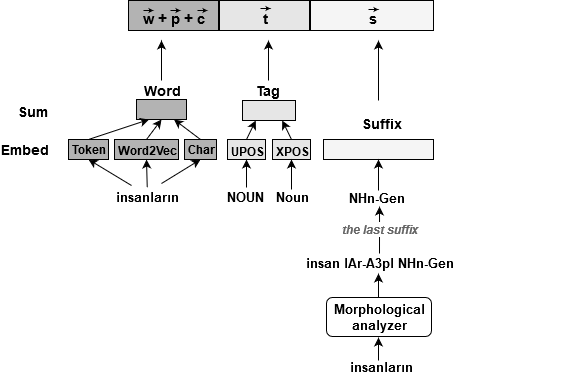}
                \caption{ Dense representation of the word \textit{insanlar{\i}n (people's)} using the Last Suffix Model. } 
            
\label{lastsuffix}
\end{figure}

Slightly different than the Inflectional Suffixes Model, here the same process is performed for only the last suffix of an input word. The vector representation of the last derivational or inflectional suffix of a word is added to the vector representation of that word. Figure \ref{lastsuffix} depicts the creation of the dense representation of the same example word in Figure \ref{infsuffix}, but this time using the Last Suffix Model. 

\subsubsection{The Suffix Vector Model}

\begin{figure}[!h]
\centering
 \includegraphics[width=1\textwidth]{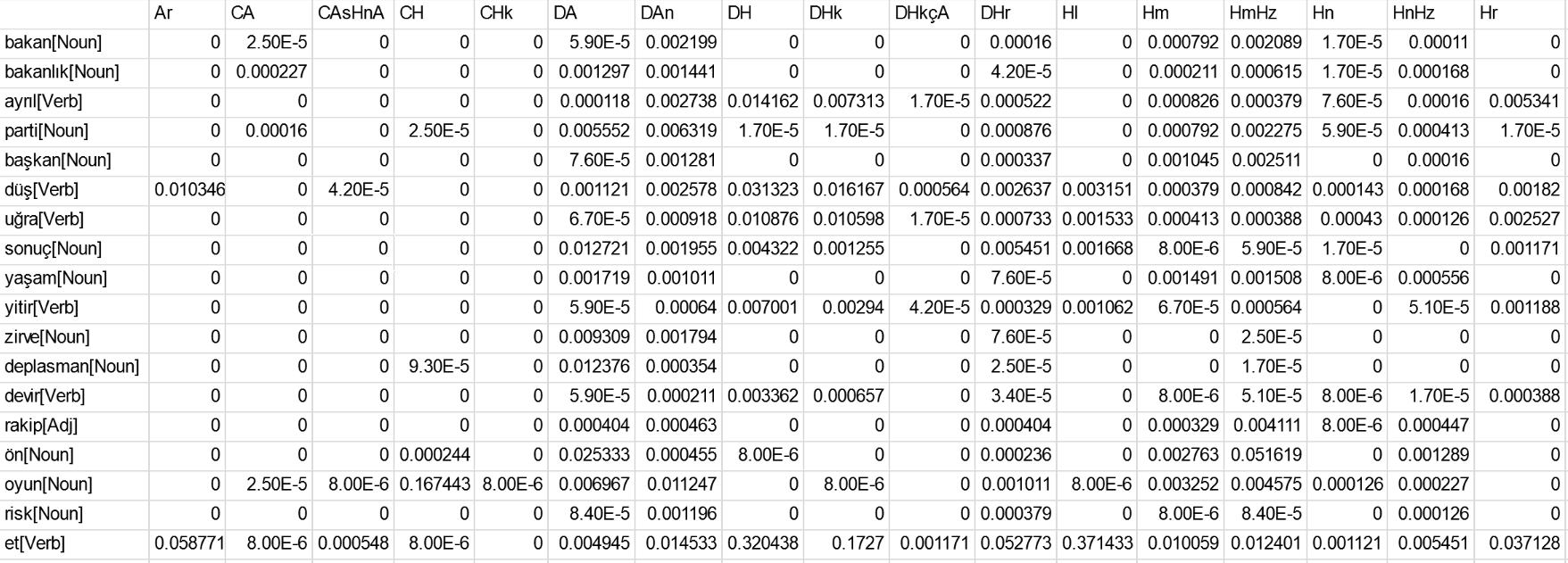}
                \caption{ A small subset of the lemma-suffix matrix created from the Newscor part of the Boun Web Corpus. } 
            
\label{lemmasufvec}
\end{figure}

This model is a bit different than the previous two models. In the Suffix Vector Model, the input words are represented through a vector of all suffixes which the lemma of that word can and cannot take. The motivation behind this model comes from the idea that the role of a word form in a sentence can be determined by considering the suffixes that the lemma of that word never takes and the suffixes it frequently takes. For instance, inflectional suffixes in Turkish indicate how the constituents of a sentence are related to each other \citep[][page 65]{goksel2005}. For this purpose, we created a lemma-suffix matrix which consists of 40K unique lemmas and 81 inflectional and derivational suffixes in Turkish. Rows of the matrix list the lemmas and columns show the normalized count of the times each lemma takes the corresponding suffix. To compute these statistics, we use Newscor part of the Boun Web Corpus \cite{sak2011resources}. Newscor is created from news documents taken from three pioneering news portals in Turkish (Milliyet, Ntvmsnbc, and Radikal) and includes 184M words in Turkish. A small subset of the lemma-suffix matrix is shown in Figure \ref{lemmasufvec} for demonstrational purposes.

\begin{figure}[!h]
\centering
 \includegraphics[width=0.6\textwidth]{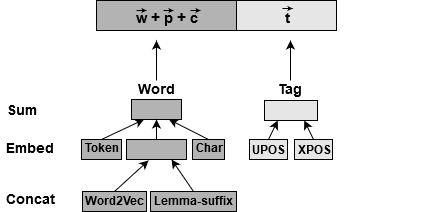}
                \caption{ Word representation model of the system with the Suffix Vector Model. } 
            
\label{suffixvec}
\end{figure}

This lemma-suffix matrix is then concatenated with the pre-trained word embedding matrix used by the parser. For each word entry in the pre-trained word embedding matrix, the row vector in the lemma-suffix matrix that belongs to the lemma of that word is found and this vector is concatenated to the end of the embedding vector of that word. If the lemma of the word does not exist in the lemma-suffix matrix, then a 81-dimensional zero-vector is concatenated instead. Figure \ref{suffixvec} depicts the word representation model of the system when we use the Suffix Vector Model.

\section{Generalization to Other Languages}

Although this study focuses on the improvement of Turkish dependency parsing, the proposed models can be adapted to other languages as well. Below we first state the adaptation of the rule-based component, which is followed by the adaptation of the morphology-based component.

\subsection{Adapting the Rule-based Parsing Approach to a Target Language}

For the rule-based parser, the AJC, AV, AJN, and NV rules can be directly applied to any language as long as the underlying structures exist in the language. 
The AC and AAJ rules use adverb lists to make decisions. They can be applied to any language by supplying the adverbs specified in the rule descriptions for that language. 

The PC rule needs a small adaptation to work for other languages because it uses the genitive and possessive marks when constructing possessive compounds. The genitive and possessive marks should be changed with the corresponding ones of the language, if applicable. 

For the CPI rule which handles complex predicates, we use a dictionary of complex predicates and idioms for Turkish. If there is a similar structure in the language the model is adapted to, supplying such a dictionary to the CPI rule will be sufficient. 
Similarly, providing the three lexicons that separate between noun compounds, possessive compounds, and reduplicated compounds for the target language will be sufficient to adapt the NC rule. 

We stated above how the rules proposed in this study can be adapted to other languages. In addition to such rule adaptations, and more importantly, effective rules for any language can be found by making a manual error analysis on the parser outputs. The grammar rules that hold for the relations between verbs, nouns, adverbs, and adjectives in a sentence can easily be applied for that language in general. The rule applications can then be integrated with the dependency parser by following the proposed strategy explained in Section \ref{intergration}.

\subsection{Adapting the Morphology-based Enhancement Approach to a Target Language}

To create our suffix-based models, we utilize a morphological analysis tool for Turkish. A similar approach can be followed for any target language with agglutinative morphology. All of the three morphology-based models can be adapted by extracting the derivational and inflectional affixes in the target language.

\section{Experiments} 

We evaluated Stanford's neural parser as the baseline system and the proposed hybrid parser with rule-based and morphology-based enhancement methods on the IMST-UD Treebank. In all of the experiments, the default set of parameters are used for the deep network that produces the parse trees.

The training part of the IMST-UD Treebank has 3,685 annotated sentences and the development and test parts have 975 annotated sentences each. We used the original partition of the treebank in our experiments. 
Unlike \citet{sulubacak2018implementing} who excluded non-projective dependencies, which are hard to predict, from the experiments, we include both projective and non-projective dependencies. It is stated in \cite{hall2010single} that more than $20\%$ of the sentences in Turkish have non-projective dependency trees. So, using only the projective sentences does not reflect the reality for Turkish. 

As in the baseline approach \cite{dozat2017stanford}, we used the Turkish word vectors from the CoNLL-17  pre-trained word vectors \cite{ginter2017conll}. In the evaluation of the dependency parsers, we used the word-based unlabeled attachment score (UAS) and the labeled attachment score (LAS) metrics, where UAS is measured as the percentage of words that are attached to the correct head, and LAS is defined as the percentage of words that are attached to the correct head with the correct dependency type. 

In our system, both the rule-based and the morphology-based methods are dependent on a morphological analyzer and disambiguator tool. For this purpose, we used the Turkish morphological analyzer and disambiguator tool by \citet{sak2011resources}. This tool takes the whole sentence as an input and analyzes and disambiguates the words with respect to their corresponding meanings in the sentence. This property is very useful for Turkish because there are many words in Turkish which have multiple morphological analyses that can be correctly disambiguated only by considering the context the word is in. The accuracy of the tool on a disambiguated Turkish corpus is reported as 96.45\% in \citet{sak2011resources}. 

Although our proposed enhancement methods use a morphological analyzer and disambiguator tool and do not rely on the gold lemmas, POS-tags, or morphological features; the baseline parser used in this study needs input sentences in CoNLL-U format where the sentence is segmented to tokens and pre-processing operations such as lemmatization and tagging is supplied for each token. Since our main aim is to improve the performance of the parser, we supplied gold tokenization and POS-tags to the parser in order to observe the pure effect of the proposed methods on parsing. However, there are many end-to-end systems for parsing from raw text that include tokenizer, tagger,
and morphological analyzer parts for the pre-processing steps. One such state-of-the-art system is the Turku Neural Parser Pipeline \cite{kanerva2018turku}. This pipeline uses the same Stanford's neural parser that we used as the baseline system and developed our integrated system on. So, our proposed parser can be directly used with the Turku pipeline or any other parsing pipeline, without relying on gold tokenization or POS-tags.

\subsection{Results}
\vspace{12pt}
\subsubsection{Ablation Study}

We made an ablation study to see how each rule contributes to the overall performance of the rule-based parsing model. We started from the baseline where no rule is applied and then add the rules one by one to the model on top of each other. At each step, we trained multiple models using different seeds for each setting and evaluated them on the development set of the IMST-UD Treebank. The attachment scores shown in the diagrams in Figures \ref{ablation1} and \ref{ablation2} are the average of the attachment scores of these multiple models. 

Table \ref{steps} shows the rules the parser uses at each step of the ablation study. In Step 5, we added AC and AAJ rules to the parser at the same time. The reason for including these rules together is that, the AC rule finds the consecutive adverbs that will possibly share the same head word and these consecutive adverb pairs are given to the AAJ and AV rules as input. When a dependency relation is constructed between an adverb and an adjective by the AAJ rule or between an adverb and a verb by the AV rule, these rules look up to the consecutive adverbs list sent by the AC rule and assign the same head to their corresponding pairs. 

\begin{table}[h!]
\caption{The ablation study steps for the rule-based parser.}
\begin{tabular}{ll}%ll|}
\hline
\bf Steps & \bf Rules \\ 
\hline
 Step 1 & No rule  \\ 
\hline
 Step 2 & CPI    \\ 
\hline
 Step 3 & CPI + NC    \\ 
\hline
 Step 4 & CPI + NC + PC    \\ 
\hline
 Step 5 & CPI + NC + PC + AC + AAJ    \\ 
\hline
 Step 6 & CPI + NC + PC + AC + AAJ + AV     \\ 
\hline
 Step 7 & CPI + NC + PC + AC + AAJ + AV  + AJC + AJN    \\ 
\hline
 Step 8 & CPI + NC + PC + AC + AAJ + AV  + AJC + AJN + NV   \\ 
\hline
\end{tabular}

\label{steps}
\end{table}

Similarly, we introduced the AJC and AJN rules to the parser at the same time in Step 7. Because, the AJC rule finds the consecutive adjective pairs that should be set as dependent words to the same head word, and the list of these consecutive adjectives are given to the AJN rule as an input. When the AJN rule forms a dependency relation between an adjective and a noun, it searches through the consecutive adjectives list and assigns the same head noun to the corresponding pairs of that adjective. 

Figure \ref{ablation1} shows the effect of each rule to the parsing performance in terms of the attachment scores. We observe that, all the rules except the AV and NV rules improve the parsing performance whereas the AV and NV rules cause a drop in both of the UAS and LAS scores. The possible reason behind this performance drop might be the over-generalizing structures of these rules. Considering the high frequency of complex sentences in Turkish which include one or more subordinate clauses other than the main clause, the risk of assigning the wrong verb as the head of an adverb is high for the AV rule. Similarly in the case of the NV rule, there is a high probability of constructing a relation between a noun and a verb falsely when there are multiple verbs in a sentence. 

So, we removed the AV and NV rules from the rule-based parser and performed the ablation study again with the new setting. The effect of the rules to the performance is depicted in Figure \ref{ablation2}. We observe that each rule now improves the parsing scores which means that none of the rules blocks the other and each of them contributes to the parsing performance of the system. 
All of the experiments on the rule-based parser were performed using this final configuration of the rules.

\begin{figure}

\begin{tikzpicture}
\begin{axis}[%
   % xlabel=rules,
%    ylabel=attachment score,
    %axis x line = bottom,axis y line = left,
    small,
    color=black,
    ytick={70,70.5,...,73},
        yticklabel style={anchor=near yticklabel,left=0.2cm},
    symbolic x coords={no rule,CPI,NC,PC,AAJ-AC,AV,AJC-AJN,NV},
    xtick=data,
    xticklabel style={rotate=60,anchor=near xticklabel,below=0.5cm},
    %,ymax=1.2 % or enlarge y limits=upper
   % title=Average Dependency Arc Length,
    nodes near coords align={vertical},
    enlargelimits=+0.15,
	ybar, % interval=0.5, 
    bar width=0.54cm,    
    legend style={font=\small, at={(0.2,0.-0.2)},anchor=north west,legend columns=2},
    ]
\addplot[color=black,fill=gray!40!white] coordinates {(no rule,70.53) (CPI,71.58) (NC,71.66) (PC,71.85) (AAJ-AC,71.98) (AV,71.45) (AJC-AJN,71.96) (NV,71.51)} node[above=3.1cm,pos=.05,black] {UAS};
\end{axis}

\begin{axis}[%
   % xlabel=rules,
%    ylabel=attachment score,
    %axis x line = bottom,axis y line = left,
    small,
    ytick={63,63.5,...,66.5},
        yticklabel style={anchor=near yticklabel,left=0.2cm},
    symbolic x coords={no rule,CPI,NC,PC,AAJ-AC,AV,AJC-AJN,NV},
    xtick=data,
    xticklabel style={rotate=60,anchor=near xticklabel,below=0.5cm},
    %,ymax=1.2 % or enlarge y limits=upper
    nodes near coords align={vertical},
    enlargelimits=+0.15,
	ybar, % interval=0.5, 
    bar width=0.54cm,    
        xshift=6cm,
        legend style={font=\small, at={(0.6,-0.1)},anchor=north east,legend columns=3},
    ]
\addplot[color=black,fill=gray!40!white] coordinates {(no rule,64.01) (CPI,65.86) (NC,66.00) (PC,66.08) (AAJ-AC,66.18) (AV,65.81) (AJC-AJN,66.06) (NV,65.69)} node[above=3.1cm,pos=.05,black] {LAS};
\end{axis}
\end{tikzpicture}
\caption{The effect of each rule to the parsing performance on the development set of the IMST-UD Treebank. Each rule is added on top of the previous rules. So, in the first step with the label {\it no rule}, there is no rule used in the model. In the second step with {\it CPI} label, the CPI rule is added to the model. In the third step with {\it NC} label, the NC rule is added to the model which means both the CPI and NC rules are present in the model etc.}
\label{ablation1}
\end{figure}

\begin{figure}

\begin{tikzpicture}
\begin{axis}[%
   % xlabel=rules,
%    ylabel=attachment score,
    %axis x line = bottom,axis y line = left,
    small,
    ytick={70.5,71,...,73},
        yticklabel style={anchor=near yticklabel,left=0.2cm},
    symbolic x coords={no rule,CPI,NC,PC,AAJ-AC,AJC-AJN},
    xtick=data,
    xticklabel style={rotate=60,anchor=near xticklabel,below=0.5cm},
    %,ymax=1.2 % or enlarge y limits=upper
   % title=Average Dependency Arc Length,
    nodes near coords align={vertical},
    enlargelimits=+0.20,
	ybar, % interval=0.5, 
    bar width=0.7cm,    
    legend style={font=\small, at={(0.2,0.-0.2)},anchor=north west,legend columns=2},
    ]
\addplot[color=black,fill=gray!40!white] coordinates {(no rule,70.53) (CPI,71.58) (NC,71.66) (PC,71.85) (AAJ-AC,71.98) (AJC-AJN,72.02)} node[above=3cm,pos=.05,black] {UAS};
\end{axis}

\begin{axis}[%
   % xlabel=rules,
%    ylabel=attachment score,
    %axis x line = bottom,axis y line = left,
    small,
    ytick={63,63.5,...,67},
    yticklabel style={anchor=near yticklabel,left=0.2cm},
    symbolic x coords={no rule,CPI,NC,PC,AAJ-AC,AJC-AJN},
    xtick=data,
    xticklabel style={rotate=60,anchor=near xticklabel,below=0.5cm},
    %,ymax=1.2 % or enlarge y limits=upper
    nodes near coords align={vertical},
    enlargelimits=+0.2,
	ybar, % interval=0.5, 
    bar width=0.7cm,    
        xshift=6cm,
        legend style={font=\small, at={(0.6,-0.1)},anchor=north east,legend columns=3},
    ]
\addplot[color=black,fill=gray!40!white] coordinates {(no rule,64.01) (CPI,65.86) (NC,66.00) (PC,66.08) (AAJ-AC,66.18) (AJC-AJN,66.39)} node[above=3cm,pos=.05,black] {LAS};
\end{axis}
\end{tikzpicture}
\caption{The effect of each rule to the parsing performance on the development set of the IMST-UD Treebank, when the AV rule and the NV rule are removed from the rule-based parsing system. Each rule is added on top of the previous rules. }
\label{ablation2}
\end{figure}

\subsection{The Experiment Results}

Table \ref{res} shows the unlabeled and labeled attachment scores of the baseline system and our proposed enhancement models on the test set of the IMST-UD Treebank.

\begin{table}%[h!]
\caption{Attachment scores of the baseline Stanford's neural parser and the proposed models on the IMST-UD Treebank.}
\begin{tabular}{lll}%ll|}
\hline
 \bf Parsing models &\multicolumn{2}{c}{\bf IMST-UD }  \\ 
\cline{2-3}
 &{\bf UAS} &{\bf LAS } \\ 
\hline
\bf {\bf baseline \cite{dozat2017stanford}} & 71.96 & 65.15  \\ 

\bf hybrid - rule & 74.03 &  67.99    \\ 

\bf hybrid - inflectional suffixes & 73.57 & 67.81   \\
\bf hybrid - last suffix & 73.95 & 68.25   \\
\bf hybrid - suffix vector & 73.09 & 66.96  \\

\bf hybrid - rule and last suffix & \bf 74.37 & \bf 68.63  \\ 
\hline
\end{tabular}

\label{res}
\end{table}

We have five different hybrid models that are built on top of the baseline model. Our first hybrid model is using the proposed rule-based approach explained in Section \ref{sec:rulebased}. The second, third, and fourth hybrid models are when we apply the corresponding versions of the morphology-based enhancement methods in Section \ref{sec:morp}. The last hybrid model is the  combination of the rule-based and morphology-based approaches. We selected the Last Suffix Model for this combination since it is the best performing one in the morphology-based methods. We combine these two models by simply concatenating their corresponding embedding vectors to the end of the original input word vector representation.

The results of the experiments show that all of our hybrid models outperform the baseline parser on the IMST-UD Treebank with p-values lower than 0.001 according to the performed randomization tests. 
We observe that the best performing model is the combination of the rule-based model and the Last Suffix Model with approximately 2.5 and 3.5 points differences in, respectively, UAS and LAS scores when compared with the baseline model.

We see that, among the three morphology-based models, the best performing one is the Last Suffix Model. The success of this model matches with the observation that the last suffix of a word determines its role in a sentence \cite{oflazer2003dependency}. This result suggests that the Last Suffix Model can accurately group the words using the last morpheme information for dependency parsing. Both of the UAS and LAS differences between this model and the other two morphology-based models are found to be statistically significant on the performed randomization test. From these results, we can conclude that the Last Suffix Model can be preferred over the other two morphology-based models with respect to the parsing performance and model simplicity. 

Although it outperforms the baseline model more than 1 point on UAS score and almost 2 points on LAS score, the Suffix Vector Model is the worst performing one among the proposed methods. Its relatively low performance can be attributed to its complex structure which includes all of the unique suffixes in Turkish. Filtering some of the suffixes by putting a frequency threshold during the construction of the lemma-suffix matrix and lowering the dimension of the vectors might improve the performance of the Suffix Vector Model.

The performance of the rule-based model significantly outperforms the Suffix Vector Model on both UAS and LAS scores. When compared with the Inflectional Suffixes Model, the rule-based model outperforms the Inflectional Suffixes Model significantly on UAS score. However, its LAS score is only slightly better than the Inflectional Suffixes Model which leads the parsing accuracy difference to be insignificant in terms of LAS score. The rule-based model performs slightly worse than the Last Suffix Model according to the LAS score and slightly better than the Last Suffix Model according to the UAS score. Both differences are small and  the performed randomization test results show that both of the differences are insignificant.

Yet, the best performance is reached when we combine the rule-based model with the Last Suffix model. The combined model outperforms all of the other models and the performance differences are found to be statistically significant. This result suggests that rule-based and morphology-based approaches improve different aspects of the deep learning-based parser on the dependency parsing of Turkish. Actually we observe that, while the rule-based model is more successful in establishing dependency relations between words, the Last Suffix Model is better at determining the relation types.

All of the p-values of the model comparisons are obtained by performing randomization tests on the model outputs and can be found in Table \ref{pval}.

\begin{table}[h!]
\caption{P-values that denote the statistical significance between the differences of parsing performances for different models. The comparisons whose p-values are equal to or higher than $0.1$ are shown with '{\bf X}' as the difference between the performances of these models are considered as not significant. }
\begin{tabular}{rrcccccc}%ll|}
\hline
&  &\bf \rotatebox[origin=l]{90}{baseline} & \bf \rotatebox[origin=l]{90}{\bf hybrid - rule} & \bf \rotatebox[origin=l]{90}{\bf hybrid - I.S.} & \bf \rotatebox[origin=l]{90}{\bf hybrid - L.S.} & \bf \rotatebox[origin=l]{90}{\bf hybrid - S.V.} & \bf \rotatebox[origin=l]{90}{\bf hybrid - rule and L.S. } \\ 
\hline
\bf  \multirow{2}{*}{\bf baseline} & \bf UAS & -  & < 0.001 & < 0.001& < 0.001& < 0.001& < 0.001 \\ 
& \bf LAS & - & < 0.001 & < 0.001& < 0.001& < 0.001& < 0.001 \\ 
\hline
\bf \multirow{2}{*}{\bf hybrid - rule}& \bf UAS &  & -  & 0.013  & \bf X & < 0.001&  0.058 \\ 
& \bf LAS &  &  - & \bf X & \bf X & < 0.001 & 0.015   \\ 
\hline
\bf \multirow{2}{*}{\bf hybrid - I.S.}& \bf UAS &  &  & - & 0.035 & 0.002& 0.001   \\ 
& \bf LAS &  &   & - & 0.02& < 0.001 & 0.004  \\ 
\hline
\bf \multirow{2}{*}{\bf hybrid - L.S.} 
& \bf UAS &  &   &  & - & < 0.001 &  0.03 \\
& \bf LAS &  &    &  & -  & < 0.001 & 0.06  \\ 
\hline
\bf \multirow{2}{*}{\bf hybrid - S.V.} 
& \bf UAS &  &   &  & & -  & < 0.001  \\ 
& \bf LAS &  &   &  & & -  & < 0.001  \\ 
\hline
\bf \multirow{2}{*}{\bf hybrid - rule and L.S.}
& \bf UAS &  &   &  & & & -  \\
& \bf LAS &  &   &  & & & -  \\ 
\hline
\end{tabular}
\label{pval}
\end{table}

A randomization test between two models is performed using the following scheme: Multiple output files of each of the two models are compared with each other using the randomization test. So, if there are 5 different output files produced using model A, and there are 5 different output files using model B, then a total of 25 p-values are calculated from the comparison of these output files. To reach a final p-value, their harmonic mean is calculated. The p-values in Table \ref{pval} are these final p-values.

We observe that the proposed methods increase the parsing accuracy of Turkish which has insufficient amount of training data. The aim of our approach is to give the parser additional information in constructing the dependency relations when learning from the training data is inadequate, i.e. there is not sufficient data to learn specific relations. The results show that both of the rule-based and morphology-based enhancements on the neural parser improve the parsing accuracy significantly. The experiments performed on Turkish suggest that the languages with rich morphology need language-specific treatments and remarkably benefit from the usage of the basic grammar rules as well as from the inclusion of morphological suffix information.

\section{Conclusions and Future Work}

We introduced a new rule-based method and three morphology-based methods for improving the accuracy of a deep learning-based parser on Turkish dependency parsing. In the rule-based approach, decisions made by the rule-based system are integrated into the word representation model of the deep learning-based parser. In the morphology-based approach, we experimented with the morphemes of the words by investigating different methods to integrate the information extracted from the morphemes into the word vector representation model of the parser. We observed that the best method of utilizing morphological information in terms of the dependency parsing of Turkish is using the last suffix of a word. A combination of the rule-based and morphology-based approaches outperforms all of the other proposed models as well as the baseline system, suggesting that Turkish dependency parsing benefits both from the linguistic grammar rules and the additional morphological information extracted from the input words. The experimental results show that our enhancement methods are useful for purely deep learning-based parsers, especially when there is not sufficient amount of training data to learn the dependency relations. The results show that the best performing model of the proposed approaches outperforms the state-of-the-art on parsing of the IMST-UD Treebank. 

As future work, we are planning to adapt our methods to other languages with restricted amount of annotated data. We believe that applying the proposed models to these languages will improve their parsing accuracies. Currently, we are working on adapting our models to Hungarian. Being an agglutinative language that employs a large system of suffixes and having an even smaller UD treebank than Turkish, we believe that Hungarian dependency parsing will benefit from our enhancement methods.

\begin{acknowledgments}
This   work   was   supported   by   the   Scientific and  Technological  Research  Council  of  Turkey (T\"{U}B\.{I}TAK) under grant number 117E971 and as BIDEB 2211 graduate scholarship. We thank Utku T\"{u}rk and Furkan Atmaca for their valuable comments and feedbacks.
\end{acknowledgments}

\hspace{12pt}
\starttwocolumn
\bibliographystyle{compling}
\bibliography{hybridparser}
\end{document}